\documentclass{article}
\usepackage{ijcai17}

\usepackage{times}
\usepackage{epsfig}
\usepackage{amsmath}
\usepackage{amssymb}
\usepackage{helvet}
\usepackage{courier}
\usepackage{multirow}
\usepackage{amsopn}
\usepackage{graphicx,subfigure}
\usepackage{array}
\usepackage{xcolor}
\usepackage{url}
\usepackage{setspace}
\usepackage{bm}
\usepackage{tabularx}
\usepackage{algorithm}
\usepackage[figuresright]{rotating}
\usepackage{CJK}
\usepackage{indentfirst}
\usepackage{epstopdf}
\usepackage{tikz}
\usetikzlibrary{bayesnet}

\title{Sharing deep generative representation for perceived image reconstruction from human brain activity}

\author{Changde Du$^{1}$, Changying Du$^{2}$, Huiguang He$^{1, 3}$\\
$^{1}$Research Center for Brain-Inspired Intelligence,\\ Institute of Automation, Chinese Academy of Sciences (CAS), Beijing, China\\
$^{2}$Laboratory of Parallel Software and Computational Science, Institute of Software, CAS, Beijing, China \\
$^{3}$Center for Excellence in Brain Science and Intelligence Technology, CAS, Shanghai,  China\\
\{duchangde2016, huiguang.he\}@ia.ac.cn, changying@iscas.ac.cn
}

\begin{document}

\maketitle

\begin{abstract}
Decoding human brain activities via functional magnetic resonance imaging (fMRI) has gained increasing attention in recent years. While encouraging results have been reported in brain states classification tasks, reconstructing the details of human visual experience still remains difficult. Two main challenges that hinder the development of effective models are the perplexing fMRI measurement noise and the high dimensionality of limited data instances. Existing methods generally suffer from one or both of these issues and yield dissatisfactory results. In this paper, we tackle this problem by casting the reconstruction of visual stimulus as the Bayesian inference of missing view in a multiview latent variable model. Sharing a common latent  representation, our joint generative model of external stimulus and brain response is not only ``deep" in extracting nonlinear features from visual images, but also powerful in capturing correlations among voxel activities of fMRI recordings. The nonlinearity and deep structure endow our model with strong representation ability, while the correlations of voxel activities are critical for suppressing noise and improving prediction.  We devise an efficient variational Bayesian method to infer the latent variables and the model parameters. To further improve the reconstruction accuracy, the latent representations of testing instances are enforced to be close to that of their neighbours from the training set  via posterior regularization. Experiments on three fMRI recording datasets demonstrate that our approach can more accurately reconstruct visual stimuli.
\end{abstract}

\section{Introduction}
Brain decoding, which aims to predict the information about external stimuli using brain activities, plays an important role in brain-machine interfaces (BMIs). Recent developments in this area have shown promising results \cite{schoenmakers2014gaussian,lee2016reconstructing}.  However, most previous researches only focus their attention on the prediction of the category of presented stimulus \cite{van2010efficient,ng2011generalized,damarla2013decoding,elahe2016brain}. Accurate reconstruction of the visual stimuli from brain activities still lacks adequate examination and requires plenty of efforts to improve. Two main challenges that hinder the development of effective models are the perplexing measurement noise of functional magnetic resonance imaging (fMRI) and the high dimensionality of limited data instances. Existing methods generally suffer from one or both of these issues and yield dissatisfactory results.

Fujiwara \emph{et al.} has proposed to use Bayesian canonical correlation analysis (BCCA) for building the reconstruction model, where image bases are automatically extracted from the measured data \cite{fujiwara2013modular}. As a latent variable model interpretation of non-probabilistic CCA,  BCCA assumes linear observation model for visual images and spherical covariance for the Gaussian distribution of voxel activities. In practice, however, linear observation model for visual images has limited representation power, and spherical covariance can not capture the correlations among voxel activities. Since the measurement noises are ubiquitous in voxel activities, utilizing the correlations among voxel activities would be critical for suppressing noise and improving prediction performance.

On the other hand, introducing deep structure into multiview representation learning is attracting more and more attentions recently \cite{wang2015deep,chandar2016correlational}.
Deep canonically correlated autoencoders (DCCAE), which consists of two deep autoencoders and optimizes the combination of canonical correlation
between the learned bottleneck representations and the reconstruction errors of the autoencoders, can extract nonlinear
features from both views and reconstruct each view by the correlational bottleneck representations \cite{wang2015deep}. Nevertheless, DCCAE did not consider the cross-reconstruction between two views, which limits its effectiveness in applications where a missing view needs to be reconstructed from the existing one.
To our knowledge, no deep multiview learning model with shared generative latent representation has been designed specifically for missing view reconstruction.

Focusing on these problems, we present a deep generative multiview model (DGMM), where we cast the reconstruction
of perceived image as the Bayesian inference of the missing view.
Sharing a common latent representation, DGMM allows us to generate visual images and fMRI activity patterns simultaneously.
For visual images, unlike BCCA, we explore nonlinear observation models parameterized by deep neural networks (DNNs), which can be  multi-layered perceptrons (MLPs) or convolutional neural networks (CNNs).  This nonlinearity and deep structure endow our model with strong representation ability.
For fMRI activity patterns, we adopt a full covariance matrix for the Gaussian distribution of voxel activities. While the full covariance matrix has the advantage of capturing the correlations among voxels, it results in severe computational issues. To reduce the complexity, we impose a low-rank assumption on the covariance matrix. This is beneficial to suppressing noise and improving prediction performance. Furthermore, we devise an efficient mean-field variational inference method to infer the latent variables and the model parameters.  To further improve the reconstruction accuracy, the latent representations of testing instances are enforced to be close to that of their neighbours from the training set via posterior regularization \cite{zhu2014bayesian}. Compared with the non-probabilistic deep multiview representation learning models mentioned above \cite{wang2015deep,chandar2016correlational}, our Bayesian model has the inherent advantage of avoiding overfitting to small training set by model averaging. Finally, extensive experimental comparisons on three fMRI recording datasets demonstrate that our approach can reconstruct visual images from fMRI measurements more accurately.

\section{Related work}

In the literature of brain decoding, there are a relatively limited number of studies reporting perceived image reconstructions to date. Miyawaki \emph{et al.} reconstructed the lower-order information such as binary contrast patterns using a combination of multi-scale local image bases whose shapes are predefined \cite{miyawaki2008visual}.  Van Gerven \emph{et al.} reconstructed handwritten digits using deep belief networks \cite{van2010neural}. Schoenmakers \emph{et al.} reconstructed handwritten characters using a straightforward linear Gaussian approach \cite{schoenmakers2013linear}. Fujiwara \emph{et al.} proposed to build the reconstruction model in which image bases can be automatically estimated by Bayesian canonical correlation analysis (BCCA) \cite{fujiwara2013modular}.  In addition, there are works trying to reconstruct movie clips \cite{nishimoto2011reconstructing,haiguang2016deep}.

Though a similar strategy to our work has been used by Fujiwara \emph{et al.} \cite{fujiwara2013modular} for visual image reconstruction, its linear observation model for visual images has limited representation power in practice. Several recently proposed deep multiview representation learning models can provide a service to visual image reconstruction \cite{wang2015deep,chandar2016correlational}. For example, deep canonically correlated autoencoders (DCCAE) with nonlinear observation models for both views has good ability to learn deep correlational representations and reconstruct each view using the learned representations respectively \cite{wang2015deep}. Compared with DCCAE, correlational neural networks (CorrNet) further considered the cross-reconstruction between two views \cite{chandar2016correlational}. However, directly applying the nonlinear maps of DCCAE and CorrNet to limited noisy brain activities is prone to overfitting.

Inspired by recent developments in deep generative models such as variational autoencoders (VAE) \cite{VAE}, we present a deep generative multiview model (DGMM), which can be viewed as a nonlinear extension of the linear method BCCA. To the best of our knowledge, this paper is the first to study visual image reconstruction via Bayesian deep learning.

\section{Perceived image reconstruction with DGMM}
In this section, we cast the reconstruction of perceived images from human brain activity as the Bayesian inference of missing view in a multiview latent
variable model.

 Assume the training set consists of paired observations from two distinct views ($\mathrm{\mathbf{X}}, \mathrm{\mathbf{Y}}$), denoted by
($\mathrm{\mathbf{x}}_1, \mathrm{\mathbf{y}}_1$)$, \ldots ,$ ($\mathrm{\mathbf{x}}_N, \mathrm{\mathbf{y}}_N$), where $N$ is the training set size, $\mathrm{\mathbf{x}}_i \in \mathbb{R}^{D_1}$ and $\mathrm{\mathbf{y}}_i \in \mathbb{R}^{D_2}$ for $i = 1, \ldots , N$. Here $\mathrm{\mathbf{X}} \in \mathbb{R}^{D_1 \times N}$ and $\mathrm{\mathbf{Y}} \in \mathbb{R}^{D_2 \times N}$ denote the visual images and fMRI activity patterns, respectively.
The presence of paired two-view data presents an opportunity to learn better representations by analyzing both views simultaneously. Therefore, we introduce the shared latent variables $\mathrm{\mathbf{Z}} \in \mathbb{R}^{K \times N}$ to relate the visual images $\mathrm{\mathbf{X}}$ to the fMRI activity patterns $\mathrm{\mathbf{Y}}$.
The shared latent variables are treated as the following Gaussian prior distribution,
\begin{equation}\label{VZ}
\begin{array}{@{ }l@{}l}
p(\mathrm{\mathbf{Z}}) = \prod_{i=1}^{N} \mathcal{N}\left(\mathbf{z}_i|\bm{0}, \mathrm{\mathbf{I}}\right).
\end{array}
\end{equation}

Since the visual image and associated fMRI activity pattern are assumed to be generated from the same latent variables, we have two likelihood functions. One is for visual images, and the other is for fMRI activity patterns.
\vskip 0.05in
\subsection{Deep generative model for perceived images}
When observation noises for image pixels are assumed to follow a Gaussian distribution with zero mean and diagonal covariance, the likelihood function of  visual images is
\begin{equation}\label{likelihoodstimuli}
\begin{array}{@{ }l@{}l}
 p_{\bm{\theta}}(\mathrm{\mathbf{X}}|\mathrm{\mathbf{Z}})\ = \ \prod_{i=1}^{N} \mathcal{N}\left(\mathbf{x}_i|\bm{\mu}_{\mathbf{x}}(\mathbf{z}_i),\ \mathrm{diag}(\bm{\sigma}^{2}_{\mathbf{x}}(\mathbf{z}_i)) \right),
\end{array}
\end{equation}
where the mean $\bm{\mu}_{\mathbf{x}}(\mathbf{z}_i)$ and covariance $\bm{\sigma}^{2}_{\mathbf{x}}(\mathbf{z}_i)$  are nonlinear functions of the latent variables $\mathbf{z}_i$. To allow for second moment of the data to be captured by the density model, we choose these nonlinear functions to be deep neural networks (DNNs), which is refer to as the $\emph{generative network}$, $g(\mathrm{\mathbf{z}})$ parameterized by $\bm{\theta}$.  Here the DNNs can be multi-layered perceptrons (MLPs) or convolutional neural networks (CNNs). Compared with linear observation model, DNNs can extract nonlinear features from visual images and capture the stages of human visual processing from early visual areas towards the ventral streams \cite{Gucclu10005,cichy2016comparison}. This nonlinearity and deep structure endow our model with strong representation ability.
\vskip 0.05in
\subsection{Generative model for fMRI activity patterns}
fMRI voxels are generally highly correlated, and the correlation can carry relevant information about stimuli or tasks, even in the absence of information in individual voxels \cite{yamashita2008sparse,hossein2016reconstruction}. However, most existing methods \cite{fujiwara2013modular,schoenmakers2013linear} simply assume a spherical or diagonal covariance for the Gaussian distribution of  voxel activities thus ignoring any correlations among voxels. Unlike them, we assume the observation noises of voxel activities follow a Gaussian distribution with zero mean and full covariance matrix. While this difference might seem minor, it is critical for the model to be able to suppress noise and improve
prediction performance.
In addition, although nonlinear transformations for fMRI activity patterns are more powerful than linear transformations (in terms of the types of features they can learn), extant multi-voxel pattern analysis (MVPA) studies have not found a clear performance benefit for nonlinear versus linear transformations. Therefore, we assume the likelihood function of fMRI activity patterns is
\begin{equation}\label{likelihoodfMRI1}
\begin{array}{@{ }l@{}l}
p(\mathrm{\mathbf{Y}}|\mathrm{\mathbf{Z}})\ = \ \prod_{i=1}^{N} \mathcal{N}\left(\mathbf{y}_i|\mathrm{\mathbf{B}}^{\top}\mathbf{z}_i, \bm{\Psi} \right).
\end{array}
\end{equation}

The model should be further complemented with priors for the projection matrix $\mathrm{\mathbf{B}}$ and the covariance matrix $\bm{\Psi} \in \mathbb{R}^{D_2 \times D_2}$. Popular  choices would be automatic relevance determination (ARD) prior and Wishart distribution for $\mathrm{\mathbf{B}}$ and $\bm{\Psi}^{-1}$, respectively,
\begin{equation}\label{ARD1}
\begin{array}{@{ }l@{}l}
 p(\bm{\tau})  = \prod_{j=1}^{D_2}\mathcal{G} \left(\tau_j|\alpha_{\tau},\beta_{\tau}\right)\\
 \\ [-5pt]
 p(\mathrm{\mathbf{B}}|\bm{\tau})  = \prod_{j=1}^{D_2} \mathcal{N}\left(\mathrm{\mathbf{b}}_j|\bm{0}, \tau_j^{-1}\mathrm{\mathbf{I}}\right)\\
 \\ [-5pt]
 p(\bm{\Psi}^{-1})  = \mathcal{W}\left(\bm{\Psi}^{-1}|\mathrm{\mathbf{V}}, n_0\right),
\end{array}
\end{equation}
where $\mathcal{G} \left(\cdot|\alpha, \beta\right)$ denotes gamma distribution with shape parameter $\alpha$ and rate parameter $\beta$, $\mathrm{\mathbf{V}}$ and $n_0$ denote the scale matrix and degrees of freedom for Wishart distribution, respectively.
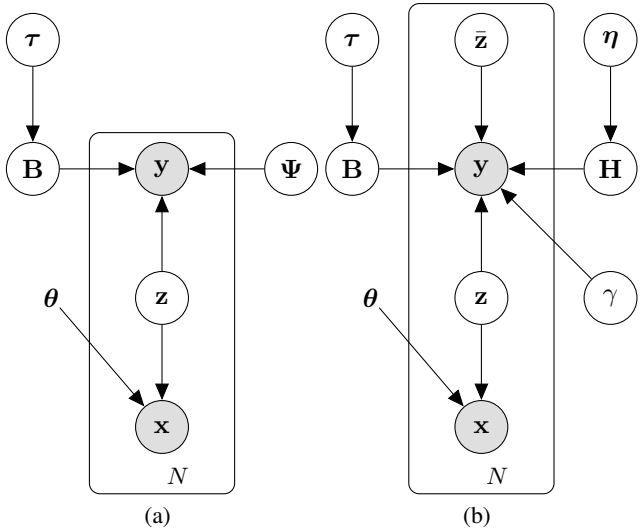
\begin{figure}[t]
  \centering
  \subfigure[] {
                \begin{tikzpicture}[scale=1, transform shape]
                \node[obs] (x1) {$\mathbf{x}$};
                \node[latent, above=of x1] (z1) {$\mathbf{z}$};
                \node[const, left=of z1] (theta1) {$\bm{\theta}$};
                \node[obs, above=of z1] (y1) {$\mathbf{y}$};
                \node[latent, left=of y1] (B1) {$\mathbf{B}$};
                \node[latent, above=of B1] (tau1) {$\bm{\tau}$};
                \node[latent, right=of y1] (Psi1) {$\mathbf{\Psi}$};

                \edge {theta1} {x1};
                \edge {z1} {x1};
                \edge {z1} {y1};
                \edge {B1} {y1};
                \edge {Psi1} {y1};
                \edge {tau1} {B1};
                \plate [xscale=2, yscale=1] {} {(x1)(y1)} {$N$} ;
                \end{tikzpicture}
                }
  \hspace {-0.25cm}
  \subfigure [] {
                \begin{tikzpicture}[scale=1, transform shape]
                \node[obs] (x1) {$\mathbf{x}$};
                \node[latent, above=of x1] (z1) {$\mathbf{z}$};
                \node[const, left=of z1] (theta1) {$\bm{\theta}$};
                \node[latent, right=of z1] (gamma1) {$\gamma$};
                \node[obs, above=of z1] (y1) {$\mathbf{y}$};
                \node[latent, above=of y1] (z_bar1) {$\bar{\mathbf{z}}$};
                \node[latent, left=of y1] (B1) {$\mathbf{B}$};
                \node[latent, right=of y1] (H1) {$\mathbf{H}$};
                \node[latent, above=of B1] (tau1) {$\bm{\tau}$};
                \node[latent, above=of H1] (eta1) {$\bm{\eta}$};

                \edge {theta1} {x1};
                \edge {z1} {x1};
                \edge {z1} {y1};
                \edge {B1} {y1};
                \edge {H1} {y1};
                \edge {z_bar1} {y1};
                \edge {gamma1} {y1};
                \edge {tau1} {B1};
                \edge {eta1} {H1};
                \plate [xscale=2, yscale=1.0] {} {(x1)(z_bar1)} {$N$} ;
                \end{tikzpicture}
                }

\caption{Graphical models for DGMM. (a) Directly uses the full covariance matrix $\bm{\Psi}$. (b) Imposing a low-rank assumption on $\bm{\Psi}$ ($\bm{\Psi} = \mathrm{\mathbf{H}^{\top}}\mathrm{\mathbf{H}} + \gamma^{-1}\mathrm{\mathbf{I}}$).}
\label{fig:graphmodel}
\end{figure}

While the above model has the advantage of capturing the correlations among voxels, it results in
severe computational issues (the cost  is cubic as a function of $D_2$). Fortunately, the problem of inferring high-dimensional covariance matrix $\bm{\Psi}$ can be solved by introducing auxiliary latent variables $\mathrm{\bar{\mathbf{Z}}} \in \mathbb{R}^{\bar{K} \times N}$ \cite{archambeau2009sparse},
\begin{equation}\label{Z_bar}
\begin{array}{@{ }l@{}l}
p(\mathrm{\bar{\mathbf{Z}}}) = \prod_{i=1}^{N} \mathcal{N}\left(\bar{\mathbf{z}}_i|\bm{0}, \mathrm{\mathbf{I}}\right),
\end{array}
\end{equation}
and rewriting the likelihood function in Eq.(\ref{likelihoodfMRI1}) as
\begin{equation}\label{likelihoodfMRI2}
\begin{array}{@{ }l@{}l}
p(\mathrm{\mathbf{Y}}|\mathrm{\mathbf{Z}}, \mathrm{\bar{\mathbf{Z}}})\ = \ \prod_{i=1}^{N} \mathcal{N}\left(\mathbf{y}_i|\mathrm{\mathbf{B}}^{\top}\mathbf{z}_i + \mathrm{\mathbf{H}}^{\top}\bar{\mathbf{z}}_i, \gamma^{-1}\mathrm{\mathbf{I}}\right),
\end{array}
\end{equation}
where ARD prior and simple gamma prior can be set for the extra projection matrix $\mathrm{\mathbf{H}} \in \mathbb{R}^{\bar{K} \times D_2}$ and variance parameter $\gamma$, respectively,
\begin{equation}\label{ARD2}
\begin{array}{@{ }l@{}l}
 p(\bm{\eta})  = \prod_{j=1}^{D_2} \mathcal{G} \left(\eta_j|\alpha_{\eta},\beta_{\eta}\right)\\
 \\ [-5pt]
 p(\mathrm{\mathbf{H}}|\bm{\eta})  = \prod_{j=1}^{D_2} \mathcal{N}\left(\mathrm{\mathbf{h}}_j|\bm{0}, \eta_j^{-1}\mathrm{\mathbf{I}}\right) \\
 \\ [-5pt]
 p(\gamma)  = \mathcal{G} \left(\gamma|\alpha_{\gamma},\beta_{\gamma}\right).
\end{array}
\end{equation}
The graphical models of DGMM are shown in Fig.\ref{fig:graphmodel}. Note that sparsity of the projection matrices $\mathrm{\mathbf{B}}$ and $\mathrm{\mathbf{H}}$ can be tuned by assigning suitable values to the hyper-parameters $(\alpha_{\tau}, \beta_{\tau})$ and $(\alpha_{\eta}, \beta_{\eta})$, respectively.

By integrating out auxiliary latent variables $\mathrm{\bar{\mathbf{Z}}}$, Eq.(\ref{likelihoodfMRI2}) can be shown to be equivalent to imposing a low-rank assumption on the covariance matrix $\bm{\Psi}$ in Eq.(\ref{likelihoodfMRI1}) ($\bm{\Psi} = \mathrm{\mathbf{H}^{\top}}\mathrm{\mathbf{H}} + \gamma^{-1}\mathrm{\mathbf{I}}$), which allows
decreasing the computational complexity. From another perspective, this low-rank assumption produces
a full factorization of the variation in fMRI data into shared components $\mathrm{\mathbf{Z}}$ and private components $\mathrm{\bar{\mathbf{Z}}}$. The ability to identify what is shared and what is non-shared makes our model be good at suppressing noise and improving prediction performance.

As short-hand notations, all hyper-parameters in the model will be denoted by $\Omega = \{\alpha_{\tau}, \beta_{\tau}, \alpha_{\eta}, \beta_{\eta}, \alpha_{\gamma}, \beta_{\gamma}\}$, while the priors by $ \Xi = \{\bm{\tau}, \bm{\eta}, \gamma \}$ and the remaining variables by $\Theta = \{\mathrm{\mathbf{B, H, Z}}, \mathrm{\bar{\mathbf{Z}}}\}$. Dependence on $\Omega$ is omitted for clarity throughout the paper. Then we can get the following posterior distribution using Bayes' rule
\begin{equation}
 p(\Theta, \Xi|\mathrm{\mathbf{X}},\mathrm{\mathbf{Y}}) = \frac{p_{\bm{\theta}}(\mathrm{\mathbf{X}}|\mathrm{\mathbf{Z}}) p(\mathrm{\mathbf{Y}}|\mathrm{\mathbf{Z}}, \mathrm{\bar{\mathbf{Z}}}) p(\Theta|\Xi)p(\Xi)} {p_{\bm{\theta}}(\mathrm{\mathbf{X,Y}})},
\end{equation}
where $p_{\bm{\theta}}(\mathrm{\mathbf{X,Y}})$ is the normalization constant.

\section{Variational posterior inference}
Given above generative model, exact inference is intractable. Here we formulate a mean-field variational approximate inference method to infer the latent variables and model parameters.
Specifically, we assume there are a family of factorable and free-form (except for $q(\mathrm{\mathbf{Z}})$) variational distributions
\begin{equation}
\begin{array}{@{ }l@{}l}
\nonumber q(\Theta, \Xi)  =  q(\mathrm{\mathbf{B}}) q(\mathrm{\mathbf{H}}) q(\mathrm{\mathbf{Z}}) q(\mathrm{\bar{\mathbf{Z}}}) q(\bm{\tau}) q(\bm{\eta}) q(\gamma),
\end{array}
\end{equation}
and define  $q(\mathrm{\mathbf{Z}})$ as a product of multivariate Gaussian distributions with diagonal covariance\footnote{We also considered to condition the posterior distribution $ q(\mathrm{\mathbf{Z}})$ on both $\mathrm{\mathbf{X}}$ and $\mathrm{\mathbf{Y}}$, but we didn't observe obvious performance improvement.}, \emph{i.e.},
\begin{equation}\label{qz}
\begin{array}{@{ }l@{}l}
\nonumber q(\mathrm{\mathbf{Z}}) = \prod\limits_{i=1}^{N} q_{\bm{\varphi}}(\mathrm{\mathbf{z}}_i|\mathrm{\mathbf{x}}_i) =  \prod\limits_{i=1}^{N} \mathcal{N}\left(\mathbf{z}_i|\bm{\mu}_{\mathbf{z}}(\mathbf{x}_i),\ \mathrm{diag}(\bm{\sigma}^{2}_{\mathbf{z}}(\mathbf{x}_i)) \right),\\
\end{array}
\end{equation}
where the mean $\bm{\mu}_{\mathbf{z}}(\mathbf{x}_i) = [{\mu_{\mathbf{z}}}_{i1}, \ldots, {\mu_{\mathbf{z}}}_{iK}]^\top$ and covariance $\mathrm{diag}(\bm{\sigma}^{2}_{\mathbf{z}}(\mathbf{x}_i)) = \mathrm{diag}({\sigma^{2}_{\mathbf{z}}}_{i1}, \ldots, {\sigma^{2}_{\mathbf{z}}}_{iK})$ are outputs of the $\emph{recognition network}$  specified by another DNN with parameters $\bm{\varphi}$.
Then the objective is to get the optimal one which minimizes the Kullback-Leibler (KL) divergence between the approximating distribution and the target posterior, \emph{i.e.},
\begin{equation}
\begin{array}{@{ }l@{}l}
\nonumber \min\limits_{q(\Theta, \Xi) \in \mathcal{P}} \mathrm{KL}\left(q(\Theta, \Xi)\| p_{\bm{\theta}}(\Theta, \Xi|\mathrm{\mathbf{X}},\mathrm{\mathbf{Y}})\right),
\end{array}
\end{equation}
where $\mathcal{P}$ is the space of probability distributions.
Equivalently, we can also bound the marginal likelihood:
\begin{align}\label{lowerboundmain}
\nonumber & \log p_{\bm{\theta}}(\mathrm{\mathbf{X,Y}}) \\
\nonumber & = \mathbb{E}_{q(\Theta, \Xi)}[ \log p_{\bm{\theta}}(\mathrm{\mathbf{X,Y}}, \Theta, \Xi) - \log q( \Theta, \Xi)] \\
\nonumber & \hspace{1.8cm}  + \mathrm{KL}\left(q(\Theta, \Xi)\| p_{\bm{\theta}}(\Theta, \Xi|\mathrm{\mathbf{X}},\mathrm{\mathbf{Y}})\right)\\
\nonumber & \geq \mathbb{E}_{q(\Theta, \Xi)}[ \log p_{\bm{\theta}}(\mathrm{\mathbf{X,Y}}, \Theta, \Xi) - \log q( \Theta, \Xi)]\\
\nonumber & =   \int q(\Theta, \Xi) [\log \frac{p(\Theta, \Xi)}{q( \Theta, \Xi)} + \log p_{\bm{\theta}}(\mathrm{\mathbf{X}} |\Theta, \Xi) \\
\nonumber & \hspace{3cm} + \log p(\mathrm{\mathbf{Y}} |\Theta, \Xi)]  d \Theta d \Xi\\
\nonumber & =  \mathcal{L}_{\mathcal{P}}(\Theta, \Xi, \mathrm{\mathbf{X, Y}}) + \mathcal{L}_{\mathcal{X}}( \Theta, \Xi, \mathrm{\mathbf{X, Y}}) + \mathcal{L}_{\mathcal{Y}}(\Theta, \Xi, \mathrm{\mathbf{X, Y}})\\
 & =  \mathcal{L}(\Theta, \Xi, \mathrm{\mathbf{X, Y}})
\end{align}
where we used the fact that KL divergence is guaranteed to be non-negative, and
\begin{equation}\nonumber
\begin{array}{@{ }l@{}l}
& \mathcal{L}_{\mathcal{P}}(\Theta, \Xi, \mathrm{\mathbf{X, Y}})  =  - D_{KL}(q(\Theta, \Xi)||p(\Theta, \Xi))\\
 \\ [-5pt]
 & \mathcal{L}_{\mathcal{X}}( \Theta, \Xi, \mathrm{\mathbf{X, Y}})  =  \mathbb{E}_{q(\Theta, \Xi)} [\log p_{\bm{\theta}}(\mathrm{\mathbf{X}}| \Theta, \Xi)]\\
  \\ [-5pt]
 & \mathcal{L}_{\mathcal{Y}}(\Theta, \Xi, \mathrm{\mathbf{X, Y}})  =  \mathbb{E}_{q(\Theta, \Xi)} [\log p(\mathrm{\mathbf{Y}}| \Theta, \Xi) ].
\end{array}
\end{equation}

Intuitively, $\mathcal{L}_{\mathcal{X}}$ and $\mathcal{L}_{\mathcal{Y}}$ can be interpreted as the (negative) expected reconstruction errors of visual images and fMRI activity patterns, respectively. Maximizing this lower bound strikes a balance between minimizing reconstruction errors of two views and minimizing the KL divergence between the approximate posterior and the prior.

\subsection{Learning $\bm{\theta}$, $\bm{\varphi}$ and $\mathrm{\mathbf{Z}}$}
Given the fixed-form approximate posterior distribution for factor $\mathrm{\mathbf{Z}}$, $\mathcal{L}_{\mathcal{P}}(\mathrm{\mathbf{Z}}, \mathrm{\mathbf{X, Y}})$ can be computed exactly as:
\begin{equation}
\begin{array}{@{ }l@{}l}
 \mathcal{L}_{\mathcal{P}}(\mathrm{\mathbf{Z}}, \mathrm{\mathbf{X, Y}})&=\sum_{i=1}^{N} - D_{KL}[q_{\bm{\varphi}}(\mathrm{\mathbf{z}}_i|\mathrm{\mathbf{x}}_i)||p(\mathrm{\mathbf{z}}_i)]\\
 \\ [-5pt]
\nonumber &  = \frac{1}{2}\sum\limits_{i=1}^{N}\sum\limits_{k=1}^{K} \left( 1 + \log ({\sigma^{2}_{\mathbf{z}}}_{ik}) - {\mu_{\mathbf{z}}}_{ik}^2 - {\sigma^{2}_{\mathbf{z}}}_{ik} \right).
\end{array}
\end{equation}

On the other hand, $\mathcal{L}_{\mathcal{X}}(\mathrm{\mathbf{Z}}, \mathrm{\mathbf{X, Y}})$ and $\mathcal{L}_{\mathcal{Y}}(\mathrm{\mathbf{Z}}, \mathrm{\mathbf{X, Y}})$ can be approximated  by Monte-Carlo sampling\cite{VAE,kingma2014semi}. Instead of sampling directly from $q_{\bm{\varphi}}(\mathrm{\mathbf{z}}_i|\mathrm{\mathbf{x}}_i)$, $\mathrm{\mathbf{z}}_i$ is
computed as a deterministic function of $\mathrm{\mathbf{x}}_i$ and some noise term such that $\mathrm{\mathbf{z}}_i$ has the desired distribution. Assuming we draw $L$ samples,  $\mathrm{\mathbf{z}}_i^{(l)}$ ($l=1, \ldots, L$) can be expressed as
\begin{equation}
\renewcommand{\arraystretch}{1.9}
\nonumber \mathrm{\mathbf{z}}_i^{(l)} = \bm{\mu}_{\mathbf{z}}(\mathbf{x}_i) + \bm{\sigma}_{\mathbf{z}}(\mathbf{x}_i) \odot \bm{\epsilon}^{(l)},
\end{equation}
where $\bm{\epsilon}^{(l)} \sim \mathcal{N}\left(\bm{0}, \mathrm{\mathbf{I}}\right)$ and $\odot$ denotes element-wise multiplication. Then the resulting Monte-Carlo approximations are
\begin{equation}
\begin{array}{@{ }l@{}l}
\mathcal{L}_{\mathcal{X}}(\mathrm{\mathbf{Z}}, \mathrm{\mathbf{X, Y}}) & =  \sum_{i=1}^{N} \{\mathbb{E}_{q(\mathrm{\mathbf{z}}_i)} \left[\log p_{\bm{\theta}}(\mathrm{\mathbf{x}}_i| \mathrm{\mathbf{z}}_i)\right]\}\\
  \\ [-5pt]
\nonumber & =  \frac{1}{L}\sum_{i=1}^{N}\sum_{l=1}^{L} \log p_{\bm{\theta}}(\mathrm{\mathbf{x}}_i| \mathrm{\mathbf{z}}_i^{(l)}),
\end{array}
\end{equation}

\begin{equation}
\hspace{-0.2cm}
\begin{array}{@{ }l@{}l}
\mathcal{L}_{\mathcal{Y}}(\mathrm{\mathbf{Z}}, \mathrm{\mathbf{X, Y}}) & =  \sum_{i=1}^{N} \{ \mathbb{E}_{q(\mathrm{\mathbf{z}}_i)} \left[\log p(\mathrm{\mathbf{y}}_i|\mathrm{\mathbf{z}}_i) \right]\}\\
  \\ [-5pt]
\nonumber & =  \frac{1}{L}\sum_{i=1}^{N}\sum_{l=1}^{L} \log p(\mathrm{\mathbf{y}}_i| \mathrm{\mathbf{z}}_i^{(l)}).
\end{array}
\end{equation}

Finally, the parameters of DNNs ($\bm{\theta}$ and $\bm{\varphi}$) can be obtained by optimizing the objective function $\mathcal{L}(\mathrm{\mathbf{Z}}, \mathrm{\mathbf{X, Y}})$ (based on minibatches) using the standard stochastic gradient based optimization methods such as SGD, RMSprop or AdaGrad \cite{duchi2011adaptive}.

\subsection{Learning $\mathrm{\mathbf{B, H}}, \mathrm{\bar{\mathbf{Z}}}$ and $\Xi$}
For a  specific factor $\pi$ (except for $\mathrm{\mathbf{Z}}$), it can be shown that when keeping all other factors fixed the optimal distribution $q^{\ast}(\pi)$ satisfies
\begin{equation}
\begin{array}{@{ }l@{}l}
\nonumber q^{\ast}(\pi)  \propto \exp\left\{\mathbb{E}_{q(\{ \Theta, \Xi \}\setminus \pi)}[\log p_{\bm{\theta}}(\mathrm{\mathbf{X,Y}}, \Theta, \Xi)]\right\}.
\end{array}
\end{equation}
For our model, thanks to the conjugacy, the resulting optimal distribution of each factor follows the same distribution as the corresponding factor.

The optimal distributions of the projection parameters can be found as a product of  multivariate Gaussian distributions:
\vspace{-0.1cm}
\begin{equation} \label{projection}
\renewcommand{\arraystretch}{1.9}
\begin{array}{@{ }l@{}l}
q^{\ast}(\mathrm{\mathbf{B}})=\prod_{j=1}^{D_2} \mathcal{N}\left(\mathrm{\mathbf{b}}_j|\bm{\mu}_{\mathrm{\mathbf{b}}_j}, \left[\langle \tau_j\rangle\mathrm{\mathbf{I}} + \langle \gamma \rangle \langle \mathrm{\mathbf{Z}}\mathrm{\mathbf{Z}}^{\top} \rangle\right]^{-1}\right)\\
q^{\ast}(\mathrm{\mathbf{H}})=\prod_{j=1}^{D_2} \mathcal{N}\left(\mathrm{\mathbf{h}}_j|\bm{\mu}_{\mathrm{\mathbf{h}}_j}, \left[\langle \eta_j\rangle\mathrm{\mathbf{I}} + \langle \gamma \rangle \langle \mathrm{\bar{\mathbf{Z}}}\mathrm{\bar{\mathbf{Z}}}^{\top} \rangle\right]^{-1}\right)
\end{array}
\end{equation}
where notation $\langle \cdot \rangle$ denotes the expectation operator, \emph{i.e.}, $\langle \pi \rangle$ means the
expectation of $ \pi $ over its current optimal distribution, and
\vspace{-0.3cm}
\begin{equation}\nonumber
\renewcommand{\arraystretch}{1.9}
\begin{array}{@{ }l@{}l}
\bm{\mu}_{\mathrm{\mathbf{b}}_j}= \bm{\Sigma}_{\mathrm{\mathbf{b}}_j} \sum_{i=1}^{N} \langle \gamma \rangle (y_{ij}-\langle\mathrm{\mathbf{h}}_j^\top \rangle \langle \mathrm{\bar{\mathbf{z}}}_i\rangle)\langle\mathrm{\mathbf{z}}_i\rangle\\
\bm{\mu}_{\mathrm{\mathbf{h}}_j}= \bm{\Sigma}_{\mathrm{\mathbf{h}}_j} \sum_{i=1}^{N} \langle \gamma \rangle (y_{ij}-\langle\mathrm{\mathbf{b}}_j^\top \rangle \langle \mathrm{\mathbf{z}}_i\rangle)\langle\mathrm{\bar{\mathbf{z}}}_i\rangle.
\end{array}
\end{equation}

The optimal distribution of the auxiliary latent variables can also be found as a product of  multivariate Gaussian distributions:
\begin{equation} \label{auxiliary latent variables}
\renewcommand{\arraystretch}{1.9}
\begin{array}{@{ }l@{\quad}l}
q^{\ast}(\mathrm{\bar{\mathbf{Z}}})=\prod_{i=1}^{N} \mathcal{N}\left(\mathrm{\bar{\mathbf{z}}}_i|\bm{\mu}_{\mathrm{\bar{\mathbf{z}}}_i}, \left[\mathrm{\mathbf{I}} + \langle \gamma \rangle \langle \mathrm{\mathbf{H}}\mathrm{\mathbf{H}}^{\top} \rangle\right]^{-1}\right)
\end{array}
\end{equation}
where
\vspace{-0.2cm}
\begin{equation}
\renewcommand{\arraystretch}{1.9}
\begin{array}{@{ }l@{\quad}l}
\nonumber \bm{\mu}_{\mathrm{\bar{\mathbf{z}}}_i}= \bm{\Sigma}_{\mathrm{\bar{\mathbf{z}}}_i} \sum_{j=1}^{D_2} \langle \gamma \rangle (y_{ij}-\langle\mathrm{\mathbf{b}}_j^\top \rangle \langle\mathrm{\mathbf{z}}_i\rangle)\langle\mathrm{\mathbf{h}}_j\rangle.
\end{array}
\end{equation}

The optimal distributions of the precision variables can be formulated as:
\begin{equation} \label{precision prior}
\renewcommand{\arraystretch}{1.9}
\begin{array}{@{ }l@{\quad}l}
q^{\ast}(\bm{\tau}) & =  \prod_{j=1}^{D_2}\mathcal{G} \left(\tau_j|\alpha_{\tau} + \frac{K}{2},\beta_{\tau} + \frac{1}{2}\langle\mathrm{\mathbf{b}}_j^\top \mathrm{\mathbf{b}}_j\rangle\right)\\
q^{\ast}(\bm{\eta}) & =  \prod_{j=1}^{D_2} \mathcal{G} \left(\eta_j|\alpha_{\eta} + \frac{\bar{K}}{2},\beta_{\eta} + \frac{1}{2}\langle\mathrm{\mathbf{h}}_j^\top \mathrm{\mathbf{h}}_j\rangle\right)\\
q^{\ast}(\gamma) & =  \mathcal{G} \left(\gamma|\alpha_{\gamma} + \frac{ND_2}{2}, \beta_{\gamma} + \frac{1}{2}\sum_{i=1}^{N}\sum_{j=1}^{D_2}\delta_{ij}^2\right)
\end{array}
\end{equation}
where $\delta_{ij}= y_{ij}- \langle\mathrm{\mathbf{b}}_j^\top\rangle \langle \mathrm{\mathbf{z}}_i\rangle - \langle\mathrm{\mathbf{h}}_j^\top \rangle \langle\mathrm{\bar{\mathbf{z}}}_i\rangle$.
\subsection{Convergence}
The inference mechanism sequentially updates the optimal distributions of the latent variables and the model parameters until convergence, which is guaranteed because the KL divergence is convex with respect to each of the factors.
\subsection{Prediction}
Using the estimated parameters, we can derive the predictive distribution for a visual image $\mathrm{\mathbf{x}}_{\mathrm{pred}}$ given a new brain activity $\mathrm{\mathbf{y}}_{\star}$.
The predictive distribution $p(\mathrm{\mathbf{x}}_{\mathrm{pred}}|\mathrm{\mathbf{y}}_{\star})$ can be formulated as follows,
\begin{equation}\label{decoding distribution}
p(\mathrm{\mathbf{x}}_{\mathrm{pred}}|\mathrm{\mathbf{y}}_{\star}) = \int p_{\bm{\theta}}(\mathrm{\mathbf{x}}_{\mathrm{pred}}|\mathrm{\mathbf{z}}_{\star}) p(\mathrm{\mathbf{z}}_{\star}|\mathrm{\mathbf{y}}_{\star}) d \mathrm{\mathbf{z}}_{\star}
\end{equation}
where the posterior distribution of latent variables $p(\mathrm{\mathbf{z}}_{\star}|\mathrm{\mathbf{y}}_{\star})$ can be derived by
\begin{eqnarray}
\nonumber p(\mathrm{\mathbf{z}}_{\star}|\mathrm{\mathbf{y}}_{\star}) = \int p(\mathrm{\mathbf{y}}_{\star}|\mathrm{\mathbf{z}}_{\star}, \mathrm{\bar{\mathbf{z}}}_{\star}, \mathrm{\mathbf{B}}, \mathrm{\mathbf{H}}, \gamma)p(\mathrm{\mathbf{z}}_{\star})p(\mathrm{\bar{\mathbf{z}}}_{\star})\\
q^{\ast}(\mathrm{\mathbf{B}})q^{\ast}(\mathrm{\mathbf{H}})q^{\ast}(\gamma) d \mathrm{\bar{\mathbf{z}}}_{\star} d \mathrm{\mathbf{B}} d \mathrm{\mathbf{H}} d \gamma
\end{eqnarray}
The posterior distribution $p(\mathrm{\mathbf{z}}_{\star}| \mathrm{\mathbf{y}}_{\star}) = p(\mathrm{\mathbf{z}}_{\star})p(\mathrm{\mathbf{y}}_{\star}| \mathrm{\mathbf{z}}_{\star})/ p(\mathrm{\mathbf{y}}_{\star})$ can be equivalently obtained by solving the following information theoretical optimization problem:
\vspace{-0.2cm}
\begin{equation} \label{KLV}
\renewcommand{\arraystretch}{1.9}
\begin{array}{@{ }l@{\quad}l}
\min\limits_{q(\mathrm{\mathbf{z}}_{\star}) \in \mathcal{P}} \mathrm{KL}\left(q(\mathrm{\mathbf{z}}_{\star})\| p(\mathrm{\mathbf{z}}_{\star}|\mathrm{\mathbf{y}}_{\star})\right)
\end{array}
\end{equation}
Expanding Eq.(\ref{KLV}) and ignoring the term unrelated to $q(\mathrm{\mathbf{z}}_{\star})$, we further get
\begin{equation} \nonumber
\renewcommand{\arraystretch}{1.9}
\begin{array}{@{ }l@{\quad}l}
& \min\limits_{q(\mathrm{\mathbf{z}}_{\star}) \in \mathcal{P}} \mathrm{KL}\left(q(\mathrm{\mathbf{z}}_{\star})\| p(\mathrm{\mathbf{z}}_{\star})\right) - \mathbb{E}_{q(\mathrm{\mathbf{z}}_{\star})}[\log p(\mathrm{\mathbf{y}}_{\star}|\mathrm{\mathbf{z}}_{\star})].
\end{array}
\end{equation}

To ensure the latent representations of testing instances are close to that of their neighbours from the training set,  we adopt the  posterior regularization\cite{zhu2014bayesian} strategy to incorporate the manifold regularization into the above posterior predictive distribution $p(\mathrm{\mathbf{z}}_{\star}|\mathrm{\mathbf{y}}_{\star})$.  Specifically, we define the following expected manifold regularization:
\begin{equation}\nonumber
\renewcommand{\arraystretch}{1.9}
\begin{array}{@{ }l@{\quad}l}
 \mathcal{R}(q(\mathrm{\mathbf{z}}_{\star}))& = \mathbb{E}_{q(\mathrm{\mathbf{z}}_{\star})}\left[\sum_{i=1}^{N} s_i \| \mathrm{\mathbf{z}}_{\star} - \mathrm{\mathbf{z}}_{i}\|^2\right]
\end{array}
\end{equation}
where $s_i$ is some similarity measure of instances $\mathrm{\mathbf{y}}_{i}$ and $\mathrm{\mathbf{y}}_{\star}$. Here we use a k-nearest neighbor graph to effectively model local geometry structure in the input space  and the affinity graph is defined as:
$$s_i=
\begin{cases}
 \exp\left(-\frac{\| \mathrm{\mathbf{y}}_{\star} - \mathrm{\mathbf{y}}_{i}\|^2}{2t^2}\right),  \ \mathrm{\mathbf{y}}_{i} \in \mathcal{N}(\mathrm{\mathbf{y}}_{\star}),\\
 0,  \ \mathrm{otherwise},
\end{cases}
$$
where $\mathcal{N}(\mathrm{\mathbf{y}}_{\star})$ denotes the k-nearest neighbors of $\mathrm{\mathbf{y}}_{\star}$.

Then our posterior regularization strategy can be formulated as
\begin{align}\label{BayesReg}
\nonumber  & \min\limits_{q(\mathrm{\mathbf{z}}_{\star}) \in \mathcal{P}} \mathrm{KL}\left(q(\mathrm{\mathbf{z}}_{\star})\| p(\mathrm{\mathbf{z}}_{\star})\right) - \mathbb{E}_{q(\mathrm{\mathbf{z}}_{\star})}[\log p(\mathrm{\mathbf{y}}_{\star}|\mathrm{\mathbf{z}}_{\star})] \\
  & \hspace{3cm} + \rho \mathcal{R}(q(\mathrm{\mathbf{z}}_{\star})),
\end{align}
where the parameter $\rho > 0$ controls the expected scale. As a direct way to impose constraints and incorporate knowledge in Bayesian models, posterior regularization is more natural and general than specially designed priors.
However, directly solving Eq.(\ref{BayesReg}) with $\mathcal{R}$ is difficult and inefficient. Let
\begin{equation}\nonumber
\renewcommand{\arraystretch}{1.9}
\begin{array}{@{ }l@{\quad}l}
h(\mathrm{\mathbf{z}}_{\star}|\rho, \mathrm{\mathbf{s}})=\exp\left\{- \rho \sum_{i=1}^{N} s_i \| \mathrm{\mathbf{z}}_{\star} - \mathrm{\mathbf{z}}_{i}\|^2\right\}
\end{array}
\end{equation}
then Eq.(\ref{BayesReg}) can be rewritten as
\begin{align} \label{BayesReg2}
\nonumber & \min\limits_{q(\mathrm{\mathbf{z}}_{\star}) \in \mathcal{P}} \mathrm{KL}\left(q(\mathrm{\mathbf{z}}_{\star})\| p(\mathrm{\mathbf{z}}_{\star})\right) - \mathbb{E}_{q(\mathrm{\mathbf{z}}_{\star})}[\log p(\mathrm{\mathbf{y}}_{\star}|\mathrm{\mathbf{z}}_{\star})]\\
& \hspace{2.3cm} - \mathbb{E}_{q(\mathrm{\mathbf{z}}_{\star})}[\log h(\mathrm{\mathbf{z}}_{\star}|\rho, \mathrm{\mathbf{s}})].
\end{align}
Solving problem Eq.(\ref{BayesReg2}), we can get the posterior distribution
\begin{align}
\nonumber  p(\mathrm{\mathbf{z}}_{\star}| \mathrm{\mathbf{y}}_{\star})  & =  \frac{p(\mathrm{\mathbf{z}}_{\star})p(\mathrm{\mathbf{y}}_{\star}| \mathrm{\mathbf{z}}_{\star})h(\mathrm{\mathbf{z}}_{\star}|\rho, \mathrm{\mathbf{s}})}{p(\mathrm{\mathbf{y}}_{\star})}\\
\nonumber  &  = \int p(\mathrm{\mathbf{y}}_{\star}|\mathrm{\mathbf{z}}_{\star}, \mathrm{\bar{\mathbf{z}}}_{\star}, \mathrm{\mathbf{B}}, \mathrm{\mathbf{H}}, \gamma)p(\mathrm{\mathbf{z}}_{\star}) h(\mathrm{\mathbf{z}}_{\star}|\rho, \mathrm{\mathbf{s}}) p(\mathrm{\bar{\mathbf{z}}}_{\star})\\
& \hspace{0.9cm} q^{\ast}(\mathrm{\mathbf{B}})q^{\ast}(\mathrm{\mathbf{H}})q^{\ast}(\gamma) d \mathrm{\bar{\mathbf{z}}}_{\star} d \mathrm{\mathbf{B}} d \mathrm{\mathbf{H}} d \gamma
\end{align}

Because the multiple integral over the random variables $\mathrm{\bar{\mathbf{z}}}_{\star}$, $\mathrm{\mathbf{B}}$, $\mathrm{\mathbf{H}}$ and $\gamma$ is intractable, we replace the random variables $\mathrm{\mathbf{B}}$, $\mathrm{\mathbf{H}}$ and $\gamma$ with the mean of estimated optimal distributions $q^{\ast}(\mathrm{\mathbf{B}}), q^{\ast}(\mathrm{\mathbf{H}})$ and $q^{\ast}(\gamma)$, respectively, to vanish the integral over $\mathrm{\mathbf{B}}$, $\mathrm{\mathbf{H}}$ and $\gamma$. Then  $p(\mathrm{\mathbf{z}}_{\star}|\mathrm{\mathbf{y}}_{\star})$ becomes
\begin{eqnarray}
p(\mathrm{\mathbf{z}}_{\star}|\mathrm{\mathbf{y}}_{\star}) =  \int p(\mathrm{\mathbf{y}}_{\star}|\mathrm{\mathbf{z}}_{\star}, \mathrm{\bar{\mathbf{z}}}_{\star}) p(\mathrm{\mathbf{z}}_{\star}) h(\mathrm{\mathbf{z}}_{\star}|\rho, \mathrm{\mathbf{s}}) p(\mathrm{\bar{\mathbf{z}}}_{\star}) d \mathrm{\bar{\mathbf{z}}}_{\star}.
\end{eqnarray}
Now the posterior distribution  $p(\mathrm{\mathbf{z}}_{\star}|\mathrm{\mathbf{y}}_{\star})$ can be found as:
\begin{equation}
\renewcommand{\arraystretch}{1.9}
 p(\mathrm{\mathbf{z}}_{\star}|\mathrm{\mathbf{y}}_{\star}) = \mathcal{N} \left(\mathrm{\mathbf{z}}_{\star}|\bm{\mu}_{\mathrm{\mathbf{z}}_{\star}}, \bm{\Sigma}_{\mathrm{\mathbf{z}}_{\star}}\right),\\
\end{equation}
where
\vspace{-0.2cm}
\begin{equation}\nonumber
\renewcommand{\arraystretch}{1.9}
\begin{array}{@{ }l@{}l}
& \bm{\Sigma}_{\mathrm{\mathbf{z}}_{\star}}  =  [ \langle\mathrm{\mathbf{B}} \mathrm{\mathbf{T}} \mathrm{\mathbf{B}}^{\top} \rangle + (1+ \rho \sum_{i=1}^{N}s_i)\mathrm{\mathbf{I}}]^{-1}\\
& \bm{\mu}_{\mathrm{\mathbf{z}}_{\star}}  =  \bm{\Sigma}_{\mathrm{\mathbf{z}}_{\star}} [ \langle \mathrm{\mathbf{B}}\rangle \mathrm{\mathbf{T}} \mathrm{\mathbf{y}}_{\star} + \rho \sum_{i=1}^{N} s_i \langle\mathrm{\mathbf{z}}_i\rangle]\\
& \mathrm{\mathbf{T}} =\gamma\mathrm{\mathbf{I}} -\gamma^2 \langle\mathrm{\mathbf{H}}^{\top}(\mathrm{\mathbf{I}} + \gamma \langle\mathrm{\mathbf{H}}\mathrm{\mathbf{H}}^{\top}\rangle)^{-1}\mathrm{\mathbf{H}}\rangle.
\end{array}
\end{equation}

However, with the likelihood of the visual image $p_{\bm{\theta}}(\mathrm{\mathbf{x}}_{\mathrm{pred}}|\mathrm{\mathbf{z}}_{\star})$ formulated by a DNN, the integral over the latent variables $\mathrm{\mathbf{z}}_{\star}$ (Eq.(\ref{decoding distribution})) can not be computed analytically. Similar as in the training phase, we can approximate this integral by Monte-Carlo sampling.
Finally, the reconstructed visual image is calculated by taking the mean of all $L$ predictions, \emph{i.e.}, $\mathrm{\mathbf{x}}_{\mathrm{pred}}=\frac{1}{L}\sum_{l=1}^{L}\mathrm{\mathbf{x}}_{\mathrm{pred}}^{(l)}$, where $\mathrm{\mathbf{x}}_{\mathrm{pred}}^{(l)}$ is the outputs of the $\emph{generative network}$, \emph{i.e.}, $\mathrm{\mathbf{x}}_{\mathrm{pred}}^{(l)}=g(\mathrm{\mathbf{z}}_{\star}^{(l)})$.

\section{Experiments}
In this section, we present extensive experimental results on fMRI recording datasets to demonstrate the effectiveness of the
proposed framework for perceived image reconstruction from human brain activity. Specifically, we compare our DGMM with the following algorithms, which use either a shallow or a deep architecture:
\begin{itemize}
\item $\mathbf{Fixed\ Bases}$ (Miyawaki \emph{et al.}): a specially designed method to reconstruct visual images by combining local image bases of multiple scales ($1\times1, 1\times2, 2\times1$, and $2\times2$ pixels covering an entire image) \cite{miyawaki2008visual}. The shapes of these predefined images bases are fixed, thus it may not be optimal for image reconstruction.
\vskip 0.05in
\item $\mathbf{Bayesian\ CCA}$ (BCCA): a probabilistic extension of CCA model that relates the fMRI activity space to the visual image space via a set of latent variables \cite{fujiwara2013modular}. BCCA assumes a linear observation model for visual images and a spherical covariance for the Gaussian distribution of fMRI voxels.
\vskip 0.05in
\item $\mathbf{Deep\ Canonically\ Correlated\ Autoencoders}$ (DCCAE): a latest deep multi-view representation learning model that consists of two autoencoders and optimizes the combination of canonical correlation between the learned bottleneck representations and the reconstruction errors of the autoencoders \cite{wang2015deep}. DCCAE do not consider the cross-reconstruction errors between two views.
\vskip 0.05in
\item $\mathbf{Deconvolutional\ Neural\ Network}$ (De-CNN): a latest neural decoding method based on multivariate linear regression and deconvolutional neural network \cite{haiguang2016deep,zeiler2011adaptive}. It is a two-stage cascade model, \emph{i.e.},  it first predicts feature-maps by multivariate linear regression, then reconstruct images by feeding the estimated feature-maps in a pre-trained deconvolutional neural network.
\end{itemize}
\subsection{Experimental testbed and setup}
\vspace{0.2cm}
\noindent \textbf{Data description}. \   We conducted experiments on three public fMRI datasets obtained from Miyawaki \emph{et al.} \cite{miyawaki2008visual}  and van Gerven \cite{van2010neural,schoenmakers2013linear}. Dataset 1, consisting of contrast-defined $10 \times 10$ patches,  contains two independent sessions \cite{miyawaki2008visual}.  One is a `random image session', in which spatially random patterns were sequentially presented.  The other is a `figure image session', in which alphabetical letters and simple geometric shapes were sequentially presented. We used fMRI data from primary visual area V1 of subject 1 (S1) for the analysis.  Note that all comparing algorithms were trained on the data from `random image session' and evaluated on the data from `figure image session'. Dataset 2 contains a hundred handwritten gray-scale digits (equal number of 6s and 9s)  at a $28 \times 28$ pixel resolution taken from the training set of the MNIST database and the fMRI data from V1, V2 and V3 \cite{van2010neural}. Dataset 3 contains 360 gray-scale handwritten characters (equal number of Bs, Rs, As, Is, Ns, and Ss) at a $56 \times 56$ pixel resolution taken from \cite{van2009new} and the fMRI data of V1, V2 taken from three subjects \cite{schoenmakers2013linear}. The visual images were downsampled from $56 \times 56$ pixels to $28 \times 28$ pixels in our experiments. The details of the 3 data sets used in our experiments had been summarized in Table \ref{benchmark data sets}.  See \cite{miyawaki2008visual,van2010neural,schoenmakers2013linear} for more information, including fMRI data acquisition and preprocessing.
\begin{table}[!htbp]
\footnotesize
\centering
\setlength\tabcolsep{2pt}
\setlength{\abovecaptionskip}{7pt}
\setlength{\belowcaptionskip}{0pt}
\qquad \qquad \qquad  \caption{The details of the 3 data sets used in our experiments.}
\label{benchmark data sets}
\begin{tabular}{|c|c|c| c|c|c|}
\hline
 Datasets           & \#Instances            &\#Pixels    &\#Voxels        & \#ROIs       & \#Training        \\ \hline\hline
 Dataset 1            & 1400                   &100        & 797            & V1           & 1320                                \\
 Dataset 2             & 100                    &784        & 3092           & V1, V2, V3   & 90                               \\
 Dataset 3         & 360                    &784        & 2420           & V1, V2       & 330                         \\ \hline
\end{tabular}
\end{table}
\vskip 0.03in
\noindent \textbf{Voxel selection}. \  Voxel selection is an important component to fMRI brain decoding because many voxels may not respond to the visual stimulus. A common approach is to choose those voxels that are maximally correlated with the visual images during training. We chose voxels for which the model provided better predictability (encoding performance). This codifies our intuition that the voxels better predicted with the visual images are those to be included in the decoding model.
The goodness-of-fit between model predictions and measured voxel activities was quantified using the coefficient of determination ($\mathrm{R}^2$) which indicates the percentage of variance that is explained by the model. In experiments, we first computed the $\mathrm{R}^2$ of each voxel using 10-fold cross-validation on training data, then voxels with positive $\mathrm{R}^2$ were selected for further analysis.

\vspace{0.2cm}
\noindent \textbf{Parameter setting}. \  The hyper-parameters of the proposed DGMM were set to $(\alpha_{\tau}, \beta_{\tau})=(\alpha_{\eta}, \beta_{\eta})=(\alpha_{\gamma}, \beta_{\gamma})=(1, 1)$ for all data sets, while 5-fold cross validation was conducted on training sets to choose better regularization parameters $\rho$ from $2^{[-8:0]}$.  For fair comparison, model parameters of other methods had also been tuned carefully. In our experiments, we considered multiple layer perceptrons (MLPs) as the type of recognition models. Inspired by the selectivity of visual areas to feature maps of varying complexity \cite{Gucclu10005,haiguang2016deep}, we set the structures of the $\emph{recognition network}$ for visual images  as `100-200', `784-256-128-10' and `784-256-128-5' for three data sets, respectively.  Specially, we considered two types of the structures for DCCAE. One has an asymmetric shape (same setup as our model for image view and a single layer setup for fMRI view, DCCAE-A), which can mimic our model in structure and function. The other one has a symmetric shape (same setup for both views, DCCAE-S), which can explore the deep nonlinear maps for fMRI data.

\subsection{Performance evaluation}
\begin{figure}[!h]
\vspace{-0.4cm}
\begin{center}
\includegraphics[height=2.5in,width=3.6in]{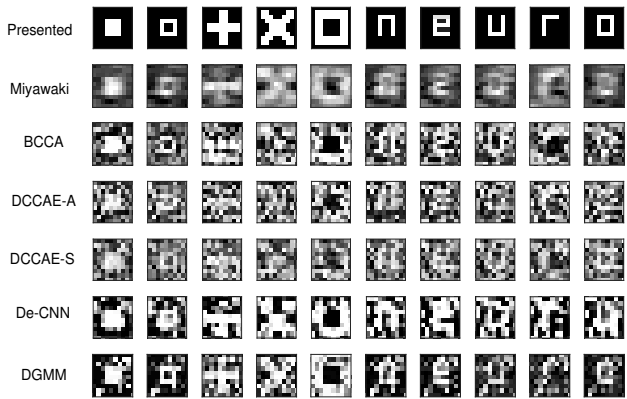}
\end{center}
\vspace{-1cm}
\caption{Image reconstructions of geometric shapes and alphabet letters taken from Dataset 1. }
\label{fig:neuro}
\end{figure}

\begin{figure}[!h]
\vspace{-0.2cm}
\begin{center}
\includegraphics[height=2.5in,width=3.6in]{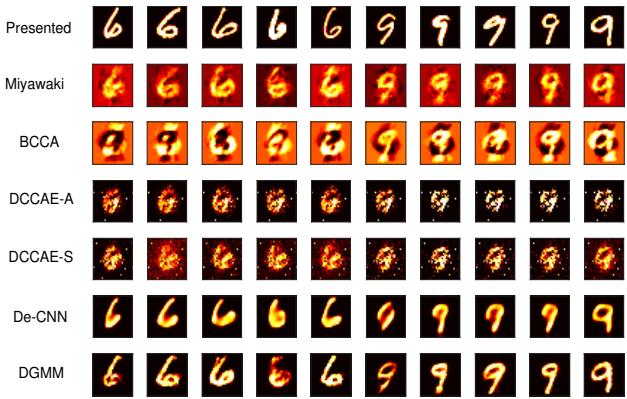}
\end{center}
\vspace{-1cm}
\caption{Image reconstructions of 10 distinct handwritten digits taken from Dataset 2.}
\label{fig:69}
\end{figure}

\begin{figure*}[!ht]
\vspace{-0.4cm}
\begin{center}
\includegraphics[height=2.8in,width=7.3in]{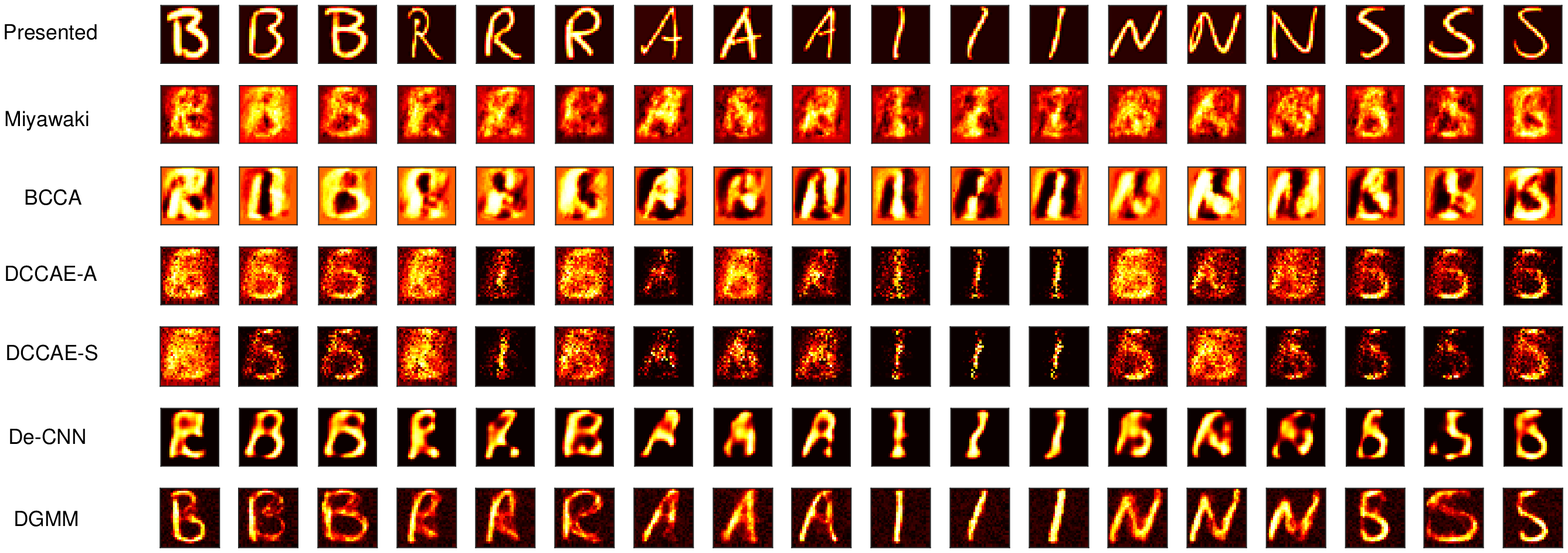}
\end{center}
\vspace{-1cm}
\caption{Examples of reconstructed 18 distinct handwritten characters taken from subject 3 of Dataset 3.}
\label{fig:brains}
\end{figure*}

The reconstructed  geometric shapes and alphabet letters, handwritten digits and handwritten characters by the proposed DGMM and other algorithms were shown in Fig.\ref{fig:neuro}, Fig.\ref{fig:69} and Fig.\ref{fig:brains}, respectively, where the first row denote presented images, and below rows are the reconstructed images obtained from all comparing algorithms.

Overall, the images reconstructed by DGMM captured the essential features of the presented images. In particular, they showed fine reconstructions for handwritten digits and characters.
Although the reconstructed geometric shapes and alphabet letters had some noise in the peripheral regions, the main shapes can be clearly distinguished. With the obtained reconstructions of handwritten digits and characters shared certain characteristics of their corresponding original images, there are subtle differences in the strokes. We attribute this phenomenon to the fact that manifold regularization imposed on the latent representations may change the details of reconstructed images. On the contrast, images reconstructed by Miyawaki's method and BCCA were coarse for all image types with noise scattered over the entire reconstructed image. Also, both DCCAE-S and DCCAE-A produced disappointing reconstructions which often lacked shapes of the presented images, especially for geometric shapes and alphabet letters. This might be due to the fact that nonlinear maps will easily over-fit the voxel activities.

To evaluate the reconstruction performance quantitatively, we used several standard image similarity metrics, including Pearson's correlation coefficient (PCC), mean squared error (MSE) and structural similarity index (SSIM) \cite{wang2004image}. Note that MSE is not highly indicative of perceived similarity, while SSIM can address this shortcoming by taking texture into account. In addition, we also performed image classification analysis to quantify the reconstruction accuracy from another perspective. Specifically, linear support vector machine (SVM) and convolutional neural network (CNN) which had been trained on the presented visual images were used as the classifiers to label the reconstructed images.  The classification accuracy of SVM (ACC-SVM) and CNN (ACC-CNN) on reconstructed images were reported. Performance comparisons were listed in Table \ref{similarity}. Note that we also listed the time consumed in training phase for all comparing algorithms in the last column for reference. Several observations can be drawn as follows.
\begin{table*}[ht]
\scriptsize
\caption{Performance of several image reconstruction methods on the test sets.  Results were averaged over 20 random seeds and all subjects (mean$\pm$std). The best performance on each dataset was highlighted.}
\centering
\renewcommand{\arraystretch}{1.28}
\setlength\tabcolsep{10pt}
\label{similarity}
\begin{tabular}{|c|l|c|c|c|c|c|c|}
  \hline
  Datasets &Algorithms & PCC & MSE & SSIM & ACC-SVM & ACC-CNN & Time(s)\\
\hline
  \hline
  \multirow{6}*{Dataset 1}
  &Miyawaki \emph{et al.}              & .609$\pm$.151            & .162$\pm$.025           & .237$\pm$.105 & $-$ & $-$  &\ 19.4$\pm$1.1\\
  &BCCA                         & .438$\pm$.215            & .253$\pm$.051           & .181$\pm$.066 & $-$ & $-$  &\ 74.9$\pm$3.0 \\
  &DCCAE-A                      & .455$\pm$.113            & .234$\pm$.029           & .166$\pm$.025 & $-$ & $-$  & 211.8$\pm$7.5 \\
  &DCCAE-S                      & .401$\pm$.100            & .240$\pm$.027           & .175$\pm$.011 & $-$ & $-$  & 254.9$\pm$9.8\\
  &De-CNN                        & .469$\pm$.149           & .263$\pm$.067           & .224$\pm$.129 & $-$ & $-$  & 108.2$\pm$2.2\\
  &DGMM                        & \textbf{.611}$\pm$.183   & \textbf{.159}$\pm$.112  & \textbf{.268}$\pm$.106  & $-$ & $-$ & 118.4$\pm$2.5\\
  \hline
  \hline
  \multirow{6}*{Dataset 2}
  &Miyawaki \emph{et al.}              & .767$\pm$.033            & .042$\pm$.007          & .466$\pm$.030   & \textbf{1.00}             &\textbf{1.00}  &\  39.9$\pm$1.2\\
  &BCCA                         & .411$\pm$.157            & .119$\pm$.017          & .192$\pm$.035   & \textbf{1.00}             &\textbf{1.00}  &\  20.7$\pm$1.0\\
  &DCCAE-A                      & .548$\pm$.044            & .074$\pm$.010          & .358$\pm$.097   & .900                      &.967$\pm$.047  &\  12.7$\pm$0.3 \\
  &DCCAE-S                      & .511$\pm$.057            & .080$\pm$.016          & .552$\pm$.088   & \textbf{1.00}             &\textbf{1.00}  &\  19.4$\pm$0.8 \\
  &De-CNN                        & .799$\pm$.062           & .038$\pm$.010           & .613$\pm$.043  & \textbf{1.00}             &\textbf{1.00}  &\  35.8$\pm$1.2 \\
  &DGMM                        & \textbf{.803}$\pm$.063   & \textbf{.037}$\pm$.014 & \textbf{.645}$\pm$.054 & \textbf{1.00}      &\textbf{1.00}  &\   18.6$\pm$1.2\\
  \hline
  \hline
  \multirow{6}*{Dataset 3}
  &Miyawaki \emph{et al.}              & .481$\pm$.096            & .067$\pm$.026          & .191$\pm$.043   & .655$\pm$.193             &.655$\pm$.113 & 128.1$\pm$4.6 \\
  &BCCA                         & .348$\pm$.138            & .128$\pm$.049          & .058$\pm$.042   & .633$\pm$.034             &.600$\pm$.098 &\  32.9$\pm$1.0 \\
  &DCCAE-A                      & .354$\pm$.167            & .073$\pm$.036          & .186$\pm$.234   & .478$\pm$.126             &.533$\pm$.072 &\  38.1$\pm$1.1  \\
  &DCCAE-S                      & .351$\pm$.153            & .086$\pm$.031          & .179$\pm$.117   & .478$\pm$.051             &.478$\pm$.155 &\  59.5$\pm$1.8 \\
  &De-CNN                        & .470$\pm$.149           & .084$\pm$.035          & .322$\pm$.118   & .589$\pm$.135             &.611$\pm$.128 &\  96.8$\pm$2.0 \\
  &DGMM                        & \textbf{.498}$\pm$.193   & \textbf{.058}$\pm$.031 & \textbf{.340}$\pm$.051  & \textbf{.767}$\pm$.115    &\textbf{.778}$\pm$.083 &\  42.4$\pm$4.2 \\
  \hline
\end{tabular}
\end{table*}

First, by comparing DGMM against the other algorithms, we can find that DGMM performs considerably better on all three data sets. In particular, the SSIM values of DGMM significantly surpass the baseline algorithms in all cases.

Second, by examining DGMM against BCCA which has a linear observation model for visual images, we can find that DGMM always outperform BCCA. This encouraging result shows that the DGMM with a DNN model for visual images is able to extract nonlinear features from visual images.

Third, DGMM shows obvious better performance than DCCAE-A and DCCAE-S. Except for ignoring cross-reconstructions, it is also caused by the fact that a linear map between voxel activities and bottleneck representation is enough to achieve good performance, while the nonlinear maps are easily overfitting under the high dimensionality of limited fMRI data instances.

Fourth, the performance of De-CNN is moderate for all data sets. We attribute this to the fact that it is a two-stage method which can't obtain the global optimal result of model parameters.

Finally, nearly $100\%$ correct classification is possible for each algorithm on Dataset 2. We believe that it is caused by the fact that digit 6 and 9 are easily to distinguish from each other. On Dataset 3, the remarkably higher classification performance on the images reconstructed by our model demonstrates the superiority of the proposed DGMM again.

\section{Conclusion and future works}

We have proposed a deep generative multiview framework to tackle the perceived image reconstruction problem.  In our framework, multiple correspondences between visual
image pixels and fMRI voxels can be found via a set of latent variables.  We also derived a predictive distribution that succeeded in reconstructing visual images from brain activity patterns. Although we focused on visual image reconstruction problem in this paper, our framework can also deal with brain encoding tasks. Extensive experimental studies have confirmed the superiority of the proposed framework.

Two challenging and promising directions can be considered in the future. First, considering the recurrent neural networks (RNNs) \cite{chung2015recurrent} in our framework, we can explore the reconstruction of dynamic vision. Second,  considering each subject's fMRI measurements as one view, we can explore  multi-subject decoding.

\section*{Acknowledgment}
This work was supported by National Natural Science Foundation of China (No. 91520202, 61602449) and Youth Innovation Promotion Association CAS.
\bibliographystyle{named}
\bibliography{egbib}
\end{document}